\documentclass[10pt, a4paper]{article}
\usepackage{lrec}
\usepackage{graphicx}
\usepackage{tabularx}
\usepackage{soul}

\usepackage{epstopdf}
\usepackage[utf8]{inputenc}
\usepackage[LGR,T1]{fontenc}
\newcommand{\textgreek}[1]{\begingroup\fontencoding{LGR}\selectfont#1\endgroup}

\usepackage{hyperref}
\usepackage{xstring}

\title{Offensive Language Identification in Greek}

\name{Zeses Pitenis\textsuperscript{1}, Marcos Zampieri\textsuperscript{2}, Tharindu Ranasinghe\textsuperscript{1}}

\address{\textsuperscript{1}University of Wolverhampton - Wolverhampton, UK \\ 
\textsuperscript{2}Rochester Institute of Technology - Rochester, NY, USA\\
         z.pitenis@wlv.ac.uk, marcos.zampieri@rit.edu, tharindu.ranasinghe@wlv.ac.uk}

\abstract{As offensive language has become a rising issue for online communities and social media platforms, researchers have been investigating ways of coping with abusive content and developing systems to detect its different types: cyberbullying, hate speech, aggression, etc. With a few notable exceptions, most research on this topic so far has dealt with English. This is mostly due to the availability of language resources for English. To address this shortcoming, this paper presents the first Greek annotated dataset for offensive language identification: the Offensive Greek Tweet Dataset (OGTD). OGTD is a manually annotated dataset containing 4,779 posts from Twitter annotated as offensive and not offensive. Along with a detailed description of the dataset, we evaluate several computational models trained and tested on this data. 
\\ \newline  }

\begin{document}

\maketitleabstract

\section{Introduction}

In the age of social media, offensive content online has become prevalent in recent years. There are many types of offensive content online such as racist and sexist posts and insults and threats targeted at individuals or groups. As such content increasingly occurs online, it has become a growing issue for online communities. This has come to the attention of social media platforms and authorities underlining the urgency to moderate and deal with such content. 
Several studies in NLP have approached offensive language identification applying machine learning and deep learning systems on annotated data to identify such content. Researchers in the field have worked with different definitions of offensive language with hate speech being the most studied among these types \cite{DBLP:journals/corr/DavidsonWMW17}. \cite{DBLP:journals/corr/WaseemDWW17} investigate the similarity between these sub-tasks. With a few noteworthy exceptions, most research so far has dealt with English, due to the availability of language resources. This gap in the literature recently started to be addressed with studies on Spanish \cite{ibereval}, Hindi \cite{Kumar2018BenchmarkingAI}, and German \cite{germeval}, to name a few. 

In this paper we contribute in this direction presenting the first Greek annotated dataset for offensive language identification: the Offensive Greek Tweet Dataset (OGTD). OGTD uses a working definition of offensive language inspired by the OLID dataset for English \cite{OLID} used in the recent OffensEval (SemEval-2019 Task 6) \cite{zampieri2019semeval}. In its version, 1.0 OGTD contains nearly 4,800 posts collected from Twitter and manually annotated by a team of volunteers, resulting in a high-quality annotated dataset. We trained a number of systems on this dataset and our best results have been obtained from a system using LSTMs and GRU with attention which achieved 0.89 F1 score.

\section{Related Work}
 
The bulk of work on detecting abusive posts online addressed particular types of such language like textual attacks and hate speech \cite{malmasi-zampieri-2017-detecting}, aggression \cite{Kumar2018BenchmarkingAI}, and others. OGTD considers a more general definition of offensiveness inspired by the first layer of the hierarchical annotation model described in \cite{OLID}. \cite{OLID} model distinguishes targeted from general profanity, and considers the target of offensive posts as indicators of potential hate speech posts (insults targeted at groups) and cyberbulling posts (insults targeted at individuals). 

\vspace{3mm}

\textbf{Offensive Language:} Previous work presented a dataset with sentences labelled as flame (i.e. attacking or containing abusive words) or okay \cite{10.1007/978-3-642-13059-5_5} with a Naïve Bayes hybrid classifier and a user offensiveness estimation using an offensive lexicon and sentence syntactic structures \cite{conf/socialcom/ChenZZX12}. A dataset of 3.3M comments from the Yahoo Finance and News website, labelled as abusive or clean, was utilized in several experiments using n-grams, linguistic and syntactic features, combined with different types of word and comment embeddings as distributional semantics features \cite{abusivenobata}. The usefulness of character n-grams for abusive language detection was explored on the same dataset with three different methods \cite{mehdad-tetreault-2016-characters}. The most recent project expanded on existing ideas for defining offensive language and presented the OLID (Offensive Language Identification Dataset), a corpus of Twitter posts hierarchically annotated on three levels, whether they contain offensive language or not, whether the offense is targeted and finally, the target of the offense \cite{OLID}. A CNN (Convolutional neural network) deep learning approach outperformed every model trained, with pre-trained FastText embeddings and updateable embeddings learned by the model as features. In OffensEval (SemEval-2019 Task 6), participants had the opportunity to use the OLID to train their own systems, with the top teams outperforming the original models trained on the dataset. 

\vspace{3mm}

\textbf{Hate Speech:} A study dataset of tweets posted after the murder of Drummer Lee Rigby in the UK, manually annotated as offensive or antagonistic in terms of race ethnicity or religion for hate speech identification with multiple classifiers \cite{doi:10.1002/poi3.85}. A logistic regression classifier trained with paragraph2vec\footnote{\href{https://github.com/thunlp/paragraph2vec}{https://github.com/thunlp/paragraph2vec}} word representations of comments from Yahoo Finance \cite{Djuric:2015:HSD:2740908.2742760}. The latest approaches in detecting hate speech include a dataset of Twitter posts, labelled as hateful, offensive or clean, used to train a logistic regression classifier with part-of-speech and word n-grams and a sentiment lexicon \cite{DBLP:journals/corr/DavidsonWMW17} and a linear SVM trained on character 4-grams, with an extra RBF SVM meta-classifier that boosts accuracy in hateful language detection \cite{DBLP:journals/corr/abs-1803-05495}. Both attempts tried to distinguish offensive language and hate speech, with the hate class being the hardest to classify.

\vspace{3mm}

\subsection{Non-English Datasets}

Research on other languages includes datasets such as: A Dutch corpus of posts from the social networking site Ask.fm for the detection of cyberbullying \cite{7010768}, a German Twitter corpus exploring the issue of hate speech targeted to refugees \cite{ross}, another Dutch corpus using data from two anti-Islamic groups in Facebook \cite{tulkens2016a}, a hate speech corpus in Italian \cite{Pelosi2017MiningOL}, an abusive language corpus in Arabic \cite{arabic}, a corpus of offensive comments from Facebook and Reddit in Danish \cite{sigurbergsson2019offensive}, another Twitter corpus in German \cite{germeval} for GermEval2018, a second Italian corpus from Facebook and Twitter \cite{DBLP:conf/evalita/BoscoDPST18}, an aggressive post corpus from Mexican Twitter in Spanish \cite{ibereval} and finally an aggressive comments corpus from Facebook in Hindi \cite{Kumar2018BenchmarkingAI}. SemEval 2019 presented a novel task: Multilingual detection of hate speech specifically against immigrants and women with a dataset from Twitter, in English and Spanish \cite{basile-etal-2019-semeval}.

\section{The OGTD Dataset}

The posts in OGTD v1.0 were collected between May and June, 2019. We used the Twitter API initially collecting tweets from popular and trending hashtags in Greece, including television programs such as series, reality and entertainment shows. Due to the municipal, regional as well as the European Parliament election taking place at the time, many hashtags included tweets discussing the elections. The intuition behind this approach is that Twitter as a microblogging service often gathers complaints and profane comments on widely viewed television and politics, and as such, this period was a good opportunity for data collection. 

Following the methodology described in \cite{OLID} and others, including a recent comparable Danish dataset \cite{sigurbergsson2019offensive}, we collected tweets using keywords such as sensitive or obscene language. Queries for tweets containing common curse words and expressions usually found in offensive messages in Greek as keywords (such as the well-known word for “asshole”, “\textgreek{μαλάκας}” (malakas) or “go to hell”, “\textgreek{στο διάολο}” (sto diaolo), etc.) returned a large number of tweets. Aiming to compile a dataset including offensive tweets of diverse types (sexist, racist, etc.) targeted at various social groups, the Twitter API was queried with expletives such as “\textgreek{πουτάνα}” (poutana, “whore”), “\textgreek{καριόλα}” (kariola, “bitch”), “\textgreek{πούστης}” (poustis, “faggot”), etc. and their plural forms, to explore the semantic and pragmatic differences of the expletives mentioned above in their different contextual environments. The challenge is to recognize between ironic and insulting uses of these swear words, a common phenomenon in Greek.   

The final query for data collection was for tweets containing “\textgreek{είσαι}” (eisai, “you are”) as a keyword, inspired by \cite{OLID}. This particular keyword is considered a stop word as it is quite common and frequent in languages but was suspected to prove helpful for building the dataset for this particular project, as offensive language often follows the following structure: auxiliary verb (be) + noun/adjective. The immediacy of social media and specifically Twitter provides the opportunity for targeted insults to be investigated, following data mining of tweets including “you are” as a keyword. In fact, many tweets present in the dataset showed users verbally insulting other users or famous people and TV personas, confirming that “\textgreek{είσαι}” was a facilitating keyword for the task in question. 

\subsection{Pre-processing and annotation} 

We collected a set of 49,154 tweets. URLs, Emojis and Emoticons were removed, while usernames and user mentions were filtered as @USER following the same methodology described in OLID \cite{OLID}. Duplicate punctuation such as question and exclamation marks was normalized. After removing duplicate tweets, the dataset was comprised of 46,218 tweets of which 5,000 were randomly sampled for annotation. We used LightTag\footnote{\href{https://www.lighttag.io/}{https://www.lighttag.io/}} to annotate the dataset due to its simple and straightforward user interface and limitless annotations, provided by the software creators.

\begin{figure}[ht]
\centering
\caption{Cohen's Kappa for each pair of annotators}
\label{fig:cohen}
\includegraphics[width=8cm]{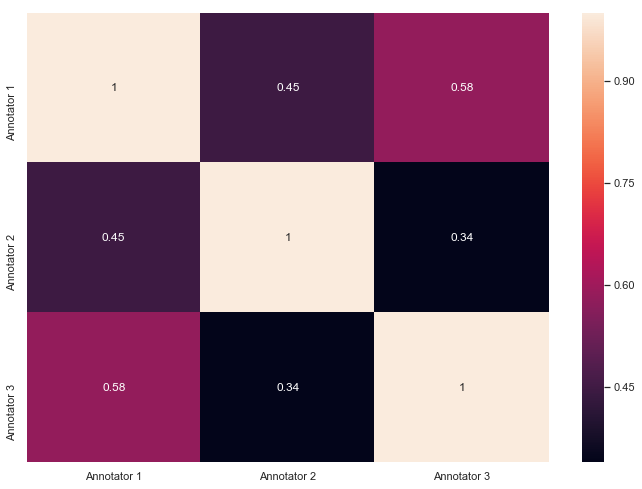}
\end{figure}

Based on explicit annotation guidelines written in Greek and our proposal of the definition of offensive language, a team of three volunteers were asked to classify each tweet found in the dataset with one of the following tags: \textit{Offensive}, \textit{Not Offensive} and \textit{Spam}, which was introduced to filter out spam from the dataset. Inter-annotator agreement was subsequently calculated and labels with 100\% agreement were deemed acceptable annotations. In cases of disagreement, labels with majority agreement above 66\% were selected as the actual annotations of the tweets in question. For labels with complete disagreement between annotators, one of the authors of this paper reviewed the tweets with two extra human judges, to get the desired majority agreement above 66\%. 
Figure \ref{fig:cohen} is a confusion matrix that shows the inter-annotator agreement or reliability, statistically measured by Cohen's kappa coefficient. The benchmark annotated dataset produced contained 4,779 tweets, containing over 29\% offensive content. The final distribution of labels in the new Offensive Greek Tweet Dataset (OGTD), along with the breakdown of the data into training and testing, is showing in Table \ref{table:1}.

\begin{table}[htb]
\centering
\begin{tabular}{lll|l}
\hline
\textbf{Labels} & \textbf{Training Set} & \textbf{Test Set} & \textbf{Total}\\ \hline
\textit{Offensive}                  & 955                   & 446   & 1,401            \\
\textit{Not Offensive}              & 2,390                  & 988   & 3,378            \\
\textit{\textbf{All}}             & 3,345                  & 1,434  & 4,779           \\ \hline
\end{tabular}
\caption{Distribution of labels in the OGTD v1.0.}
\label{table:1}
\end{table}

\section{Methods}

Before experimenting with OGTD, an unique aspect of Greek which is the accentuation of characters for correct pronunciation needed to be normalized. When posting a tweet, many users omit accents due to their haste, resulting in a mixed dataset containing fully accented tweets, partially-accented tweets, and non-accented tweets. To achieve data uniformity and to avoid ambiguity, every word is lower-cased and then normalized to its non-accented equivalent. 

Several experiments were conducted with the OGTD, each one utilizing a different combination from a pool of features (e.g. TF/IDF unigrams, bigrams, POS and dependency relation tags) to train machine learning models. These features were selected based on previous methodology used by researchers and taking the dataset size into consideration. The TF-IDF weighted features are often used for text classification and are useful for determining how important a word is to a post in a corpus. The threshold for corpus specific words was set to 80\%, ignoring terms appearing in more than 80\% of the documents while the minimum document frequency was set to 6, and both unigrams and bigrams were tested. Given the consistent use of linguistic features for training machine learning models and results from previous work for offensive language detection, part-of-speech (POS) and dependency relation tags were considered as additional features. Using the spaCy\footnote{\href{https://spacy.io/}{https://spacy.io/}} pipeline for Greek, POS-tags and dependency relations were extracted for every token in a tweet, which were then transformed to count matrices. A sentiment lexicon was considered, but one suitable for this project is as of yet unavailable for Greek.

\begin{table*}[htb]
\centering
\begin{tabular}{l|ccc|ccc|ccc|c}

\hline
                                     & \multicolumn{3}{c|}{\textbf{Not Offensive}} & \multicolumn{3}{c|}{\textbf{Offensive}}             & \multicolumn{3}{c|}{\textbf{Weighted Average}}      & \textbf{}         \\ \hline
\multicolumn{1}{l|}{\textbf{Model}} & \textbf{P}   & \textbf{R}   & \textbf{F1}   & \textbf{P} & \textbf{R} & \textbf{F1}               & \textbf{P} & \textbf{R} & \textbf{F1}               & \textbf{F1 Macro} \\ \hline
\textit{Linear SVM}                  & 0.83         & 0.98         & 0.90          & 0.92       & 0.57       & 0.70 & 0.86       & 0.85       & 0.84 & \textbf{0.80}     \\
\textit{RBF SVM}                     & 0.76         & 0.99         & 0.86          & 0.96       & 0.31       & 0.47 & 0.82       & 0.78       & 0.74 & 0.66              \\
\textit{SGDC}                        & 0.84         & 0.96         & 0.90          & 0.86       & 0.61       & 0.71 & 0.85       & 0.85       & 0.84 & \textbf{0.80}     \\
\textit{Multinomial NB}              & 0.77         & 0.99         & 0.86          & 0.94       & 0.33       & 0.49 & 0.82       & 0.78       & 0.75 & 0.67 
   \\
\textit{Bernoulli NB}                & 0.83         & 0.89         & 0.86          & 0.71       & 0.61       & 0.66 & 0.80       & 0.80       & 0.80 & 0.76              \\ \hline
\end{tabular}
\caption[Results with TF/IDF unigram features]{Results for offensive language detection with TF/IDF unigram features. For each model, Precision (P), Recall (R), and F1 are reported on all classes, and weighted averages. Macro-F1 is also listed (best in bold).}
\label{table:6}
\end{table*}

\begin{table*}[htb!]
\centering
\begin{tabular}{l|ccc|ccc|ccc|c}
\hline
                                     & \multicolumn{3}{c|}{\textbf{Not Offensive}} & \multicolumn{3}{c|}{\textbf{Offensive}}             & \multicolumn{3}{c|}{\textbf{Weighted Average}}      & \textbf{}         \\ \hline
\multicolumn{1}{l|}{\textbf{Model}} & \textbf{P}   & \textbf{R}   & \textbf{F1}   & \textbf{P} & \textbf{R} & \textbf{F1}               & \textbf{P} & \textbf{R} & \textbf{F1}               & \textbf{F1 Macro} \\ \hline
\textit{Linear SVM}                  & 0.82         & 0.98         & 0.90          & 0.92       & 0.54       & 0.68 & 0.86       & 0.84       & 0.83 & \textbf{0.79}     \\
\textit{RBF SVM}                     & 0.74         & 1.00         & 0.85          & 0.98       & 0.24       & 0.39 & 0.82       & 0.76       & 0.71 & 0.62              \\
\textit{SGDC}                        & 0.84         & 0.94         & 0.89          & 0.81       & 0.61       & 0.69 & 0.83       & 0.83       & 0.83 & \textbf{0.79}     \\
\textit{Multinomial NB}              & 0.77         & 0.99         & 0.87          & 0.93       & 0.32       & 0.48 & 0.82       & 0.79       & 0.75 & 0.67              \\
\textit{Bernoulli NB}                & 0.82         & 0.88         & 0.85          & 0.68       & 0.57       & 0.62 & 0.78       & 0.79       & 0.78 & 0.74              \\ \hline
\end{tabular}
\caption[Results with TF/IDF bigram features]{Results for offensive language detection with TF/IDF bigram features. For each model, Precision (P), Recall (R), and F1 are reported on all classes, and weighted averages. Macro-F1 is also listed (best in bold).}
\label{table:7}
\end{table*}

\begin{table*}[htb!]
\centering
\begin{tabular}{l|ccc|ccc|ccc|c}
\hline
                                     & \multicolumn{3}{c|}{\textbf{Not Offensive}} & \multicolumn{3}{c|}{\textbf{Offensive}}             & \multicolumn{3}{c|}{\textbf{Weighted Average}}      & \textbf{}         \\ \hline
\multicolumn{1}{l|}{\textbf{Model}} & \textbf{P}   & \textbf{R}   & \textbf{F1}   & \textbf{P} & \textbf{R} & \textbf{F1}               & \textbf{P} & \textbf{R} & \textbf{F1}               & \textbf{F1 Macro} \\ \hline
\textit{Linear SVM}                  & 0.84         & 0.96         & 0.90          & 0.88       & 0.58       & 0.70 & 0.85       & 0.85       & 0.83 & \textbf{0.80}     \\
\textit{SGDC}                        & 0.80         & 0.95         & 0.87          & 0.81       & 0.48       & 0.61 & 0.81       & 0.80       & 0.79 & 0.74              \\
\textit{Multinomial NB}              & 0.77         & 0.95         & 0.85          & 0.78       & 0.36       & 0.49 & 0.77       & 0.77       & 0.74 & 0.67              \\
\textit{Bernoulli NB}                & 0.80         & 0.78         & 0.79          & 0.54       & 0.58       & 0.56 & 0.72       & 0.72       & 0.72 & 0.68              \\ \hline
\end{tabular}
\caption[Results with TF/IDF unigram features, POS and dependency relation tags.]{Results for offensive language detection with TF/IDF unigram features, POS and dependency relation tags. For each model, Precision (P), Recall (R), and F1 are reported on all classes, and weighted averages. Macro-F1 is also listed (best in bold).}
\label{table:8}
\end{table*}

\begin{table*}[htb!]
\centering
\begin{tabular}{l|ccc|ccc|ccc|c}
\hline
                                     & \multicolumn{3}{c|}{\textbf{Not Offensive}} & \multicolumn{3}{c|}{\textbf{Offensive}} & \multicolumn{3}{c|}{\textbf{Weighted Average}} & \textbf{}         \\ \hline
\multicolumn{1}{l|}{\textbf{Model}} & \textbf{P}   & \textbf{R}   & \textbf{F1}   & \textbf{P}  & \textbf{R}  & \textbf{F1} & \textbf{P}    & \textbf{R}    & \textbf{F1}    & \textbf{F1 Macro} \\ \hline
\textit{Linear SVM}                  & 0.84         & 0.97         & 0.90          & 0.91        & 0.58        & 0.71        & 0.86          & 0.85          & 0.84           & \textbf{0.80}     \\
\textit{SGDC}                        & 0.74         & 0.99         & 0.85          & 0.93        & 0.22        & 0.35        & 0.80          & 0.75          & 0.69           & 0.60              \\
\textit{Multinomial NB}              & 0.77         & 0.99         & 0.86          & 0.93        & 0.33        & 0.49        & 0.82          & 0.78          & 0.75           & 0.68              \\
\textit{Bernoulli NB}                & 0.83         & 0.86         & 0.84          & 0.66        & 0.61        & 0.63        & 0.78          & 0.78          & 0.78          & 0.74              \\ \hline
\end{tabular}
\caption[Results for offensive language detection with TF/IDF unigram features and POS tags]{Results for offensive language detection with TF/IDF unigram features and POS tags. For each model, Precision (P), Recall (R), and F1 are reported on all classes, and weighted averages. Macro-F1 is also listed (best in bold).}
\label{table:9}
\end{table*}

\begin{table*}[htb!]
\centering
\begin{tabular}{l|ccc|ccc|ccc|c}
\hline
                                     & \multicolumn{3}{c|}{\textbf{Not Offensive}} & \multicolumn{3}{c|}{\textbf{Offensive}} & \multicolumn{3}{c|}{\textbf{Weighted Average}} & \textbf{}         \\ \hline
\multicolumn{1}{l|}{\textbf{Model}} & \textbf{P}   & \textbf{R}   & \textbf{F1}   & \textbf{P}  & \textbf{R}  & \textbf{F1} & \textbf{P}    & \textbf{R}    & \textbf{F1}    & \textbf{F1 Macro} \\ \hline
\textit{Linear SVM}                  & 0.84         & 0.97         & 0.90          & 0.90        & 0.58        & 0.70        & 0.86          & 0.85          & 0.84           & \textbf{0.80}     \\
\textit{SGDC}                        & 0.87         & 0.66         & 0.75          & 0.51        & 0.78        & 0.61        & 0.76          & 0.70          & 0.71           & 0.68              \\
\textit{Multinomial NB}              & 0.77         & 0.97         & 0.86          & 0.85        & 0.35        & 0.49        & 0.79          & 0.78          & 0.74           & 0.67              \\
\textit{Bernoulli NB}                & 0.82         & 0.81         & 0.81          & 0.58        & 0.60        & 0.59        & 0.74          & 0.74          & 0.74           & 0.70              \\ \hline
\end{tabular}
\caption[Results with TF/IDF unigram features and dependency relation tags]{Results for offensive language detection with TF/IDF unigram features and dependency relation tags. For each model, Precision (P), Recall (R), and F1 are reported on all classes, and weighted averages. Macro-F1 is also listed (best in bold).}
\label{table:10}
\end{table*}

For the first six deep learning models we used Greek word embeddings trained on a large Greek web corpus \cite{Outsios2018WordEF}. Each Greek word can be represented with a 300 dimensional vector using the trained model. The vector then can be used to feed in to the deep learning models which will be described in section \ref{sec:deep}. For the last deep learning architecture we wanted to use a BERT \cite{Devlin2019BERTPO} model trained on Greek. However there was no BERT model available for Greek language. The model that came closest our requirement was multilingual BERT model \footnote{\href{https://github.com/google-research/bert} {https://github.com/google-research/bert}} trained on 108 languages \cite{Devlin2019BERTPO} including Greek. Since training BERT is a very computationaly expensive task we used the available multilingual BERT cased model for the sixth deep learning architecture. 

\subsection{Models}

\subsubsection{Classical Machine Learning Models}
Every classical model was considered on the condition it could take matrices as input for fitting and was trained with the default settings because of the size of the dataset. Five models were trained: Two SVMs, one with linear kernel and the other with a radial basis function kernel (RBF), both with a value of 1 in the penalty parameter C of the error term. The gamma value of the RBF SVM which indicates how much influence a single training example has, was set to 2. The third classifier trained was another linear classifier with Stochastic Gradient Descent (SGDC) learning. The gradient of the loss is estimated each sample at a time and the SGDC is updated along the way with a decreasing learning rate. The parameters for maximum epochs and the stopping criterion were defined using the default values in scikit-learn. The final classifier was two models based on the Bayes theorem: Multinomial Naïve Bayes, which works with occurrence counts, and Bernoulli Naïve Bayes, which is designed for binary features.

\subsubsection{Deep Learning Models}
\label{sec:deep}
Six different deep learning models were considered. All of these models have been used in an aggression detection task. The models are Pooled GRU \cite{plum2019rgcl1},  Stacked  LSTM  with  Attention \cite{plum2019rgcl1},  LSTM and GRU with Attention \cite{plum2019rgcl1},  2D Convolution with Pooling \cite{ranasinghe2019brums},  GRU with Capsule \cite{Hettiarachchi:RANLP:2019},  LSTM with Capsule and Attention \cite{ranasinghe2019brums} and BERT \cite{Devlin2019BERTPO}. These models has been used in HASOC 2019 and achieved a third place finish in English task and a eighth place finish in German and Hindi subtasks \cite{ranasinghe2019brums}. Parameters described in \cite{ranasinghe2019brums} were used as the default parameters in order to ease the training process. The code for the deep learning has been made available on Github \footnote{\href{https://github.com/tharindudr/aggression-detection-greek} {https://github.com/tharindudr/aggression-detection-greek}}.  

\begin{table*}[htb!]
\centering
\begin{tabular}{l|ccc|ccc|ccc|c}
\hline
                                     & \multicolumn{3}{c|}{\textbf{Not Offensive}} & \multicolumn{3}{c|}{\textbf{Offensive}} & \multicolumn{3}{c|}{\textbf{Weighted Average}} & \textbf{}         \\ \hline
\multicolumn{1}{l|}{\textbf{Model}}      & \textbf{P}   & \textbf{R}   & \textbf{F1}   & \textbf{P}  & \textbf{R}  & \textbf{F1} & \textbf{P}    & \textbf{R}    & \textbf{F1}    & \textbf{F1 Macro} \\ \hline
\textit{Pooled GRU}                      & 0.90         & 0.99         & 0.95          & 0.94        & 0.65        & 0.75        & 0.91          & 0.86          & 0.86           & 0.87             \\
\textit{Stacked LSTM with Attention}     & 0.91         & 0.99         & 0.96          & 0.95        & 0.66        & 0.76        & 0.92          & 0.87          & 0.87           & 0.88              \\
\textit{LSTM and GRU with Attention}     & 0.92         & 0.99         & 0.96          & 0.96        & 0.68        & 0.77        & 0.93          & 0.88          & 0.88           & \textbf{0.89}     \\
\textit{2D Convolution with Pooling}     & 0.91         & 0.98         & 0.96          & 0.95        & 0.64        & 0.74        & 0.90          & 0.86          & 0.85           & 0.88              \\ 
\textit{GRU with Capsule}                & 0.92         & 0.99         & 0.95          & 0.94        & 0.64        & 0.75        & 0.91          & 0.86          & 0.85           & 0.88              \\ 
\textit{LSTM with Capsule and Attention} & 0.91         & 0.98         & 0.95          & 0.94        & 0.66        & 0.75        & 0.90          & 0.86          & 0.86           & 0.87              \\
\textit{BERT-Base Multilingual Cased}                            & 0.85         & 0.84         & 0.84          & 0.65        & 0.60        & 0.58        & 0.77          & 0.76          & 0.75           & 0.73              \\\hline
\end{tabular}
\caption[Results for Deep Learning Models with Greek Word Embeddings]{Results for offensive language detection for Deep Learning models with Greek word embeddings. For each model, Precision (P), Recall (R), and F1 are reported on all classes, and weighted averages. Macro-F1 is also listed (best in bold).}
\label{table:11}
\end{table*}

\subsection{Results}

The performance of individual classifiers for offensive language identification with TF/IDF unigram features is demonstrated in table \ref{table:6} below. We can see that both linear classifiers (SVM and SGDC) outperform the other classifiers in terms of macro-F1, which does not take label imbalance into account. The Linear SVM and SGDC perform almost identically, with the Linear SVM performing slightly better in recall score for the \textit{Not Offensive} class and SGDC in recall score for the \textit{Offensive} class. Bernoulli Naïve Bayes performs better than all classifiers in recall score for the \textit{Offensive} class but yields the lowest precision score of all classifiers. While the RBF SVM and Multinomial Naïve Bayes yield better recall score for the \textit{Not Offensive} class, their recall scores for the \textit{Offensive} class are really low. For a binary text classification task like offensive language detection, a high recall score for both classes, especially for the \textit{Offensive} class, is important for a model to be considered successful. Thus, the Linear SVM can be considered the marginally best model trained with OGTD, as its weighted average precision and recall scores are higher. 

Models trained with TF/IDF bigram features performed worse, with scores of all evaluation metrics dropping with the exception of Multinomial Naïve Bayes which improved in F1-score for the \textit{Not Offensive} class. The full results are reported in table \ref{table:7} below. Three other approaches were opted for training the models with the implementation of POS and dependency relation tags via a transformation pipeline, also including TF/IDF unigram features, performing better than the addition of bigrams.

Experiments with linguistic features were conducted, to inspect their efficiency for this task. For these experiments, the RBF SVM was not used due to data handling problems by the model in the scikit-learn library. In the first experiment, TF/IDF unigram features were combined with POS and dependency relation tags. The results of implementing all three features are shown in table \ref{table:8} below. While the Linear SVM model improved the recall score on the previous model trained with bigrams, the other models show a significant drop in their performance. 

In the next experiment, POS tags were used in conjunction with TF/IDF unigram features. Surprisingly, the addition of POS tags in the Linear SVM yields the same F1-score  as the first model trained on TF/IDF unigram features, yielding lower precision scores for both classes, while the recall score for the \textit{Offensive} class improved marginally. The Naïve Bayes models show a marginal decrease in their performance. On the other hand, the performance of SGDC significantly decreases with POS tags only and, interestingly enough, its recall score for the \textit{Offensive} class is the worst among classifiers. The complete results are presented in table \ref{table:9} below. 

The experiment with linguistic features was the combination of dependency relation tags with TF/IDF unigrams. This experimented yielded the same F1-score of 80\% as the other Linear SVM classifiers, performing almost identically with the previous model trained with POS tags, only bested in precision for the \textit{Offensive} class. While the recall score for \textit{Offensive} instances improves on the first model trained only on TF/IDF unigrams by 0.01\%, the recall score for \textit{Not Offensive} instances drops by the same amount. The recall score for the \textit{Not Offensive} class was already high, so this increase in recall score could slightly facilitate the offensive language detection task. Without improving upon the first SGDC presented, the SGDC rised in performance overall and as for the Naïve Bayes representatives, the both the Multinomial and Bernoulli approaches performed better than in the second experiment. The complete results are shown in table \ref{table:10} below.

The performance of the deep learning models is presented in table \ref{table:11}. As we can see \textit{LSTM and GRU with Attention} outperformed all the other models in-terms of macro-f1. Notably it outperformed all other classifical models and deep learning models in precision, recall and f1 for \textit{Offensive} class as well as the \textit{Not Offensive} class. However, fine tuning BERT-Base Multilingual Cased model did not achieve good results. For this task monolingual Greek word embeddings perform significantly better than the multilingual bert embeddings. \textit{LSTM and GRU with Attention} can be considered as the best model trained for OGTD.

\subsection{Discussion}

The data annotated in OGTD proved to be facilitating in offensive language detection with a significant success for Greek, taking into consideration its size and label distribution, with the best model (LSTM and GRU with Attention) achieving a F1-macro of 0.89. Among the classical machine learning approaches, the linear SVM model achieved the best results, 0.80, whereas the the Stochastic Gradient Descent (SGD) learning classifier yielded the best recall score for the \textit{Offensive} class, at 0.61. In terms of features used, TF/IDF matrices of word unigrams proved to work work well with multiple classical ML classifiers. Overall, it is clear that deep learning models with word embedding feature provide better results than the classical machine learning models.

Of the linguistic features, POS tags improved the performance of the Linear SVM marginally in terms of recall for the \textit{Offensive} class, other classifiers deteriorated in their performance.It is not yet clear whether this is due to the accuracy of the Greek model available for spaCy in producing such tags or the tags themselves as features and is a subject that can be explored with further improvements of spaCy or other NLP tools developed for Greek. The dataset itself contains many instances with neologisms, creative uses of language or and even rare slang words, therefore training the existing model with such instances could improve both spaCy's accuracy for POS and dependency relation tags and the Linear SVM's performance in text classification for Greek.


\section{Conclusion}

This paper presented the Offensive Greek Tweet Dataset (OGTD), a manually annotated dataset for offensive language identification and the first Greek dataset of its kind. The OGTD v1.0 contains a total of 4,779 tweets, encompassing posts related to an array of topics popular among Greek people (e.g. political elections, TV shows, etc.). Tweets were manually annotated by a team volunteers through an annotation platform. We used the same guidelines used in the annotation of the English OLID dataset \cite{OLID}. Finally, we run several machine learning and deep learning classifiers and the best results were achieved by a LSTM and GRU with Attention model.

\subsection{Ongoing - OGTD v2.0 and OffensEval 2020}

We have recently released OGTD v2.0 as training data for OffensEval 2020 (SemEval-2020 Task 12) \cite{zampieri-etal-2020-semeval}.\footnote{https://sites.google.com/site/offensevalsharedtask/home} The reasoning behind the expansion of the dataset was to have a larger Greek dataset for the competition. New posts were collected in November 2019 following the same approach we used to compile v1.0 described in this paper. This second batch of tweets included tweets with trending hashtags, shows and topics from Greece at the time. Additionally, keywords that proved to retrieve interesting tweets in the first version were once again used in the search, along with new keywords like pejorative terms. When the collection was finished, 5,508 tweets were randomly sampled to be then annotated by a team of volunteers. The annotation guidelines were the same ones we used for v1.0. OGTD v2.0 combines the existing with the newly annotated tweets in a larger dataset of 10,287 instances.

\begin{table}[htb]
\centering
\begin{tabular}{lll|l}
\hline
\textbf{Labels} & \textbf{Training Set} & \textbf{Test Set} & \textbf{Total}\\ \hline
\textit{Offensive}                  & 2,486                   & 425   & 2,911            \\
\textit{Not Offensive}              & 6,257                  & 1119   & 7,376            \\
\textit{\textbf{All}}             & 8,743                  & 1,544  & 10,287           \\ \hline
\end{tabular}
\caption{Distribution of labels in the OGTD v2.0.}
\label{table:12}
\end{table}

Finally, both OGTD v1.0 and v2.0 provide the opportunity for researchers to test cross-lingual learning methods as it can be used in conjunction with the English OLID and other datasets annotated using the same guidelines such as the one by \newcite{sigurbergsson2019offensive} for Danish and by \newcite{coltekikin2020} for Turkish while simultaneously facilitating the development of language resources for NLP in Greek.

\section*{Acknowledgements}
We would like to acknowledge Maria, Raphael and Anastasia, the team of volunteer annotators that provided their free time and efforts to help us produce v1.0 of the dataset of Greek tweets for offensive language detection, as well as Fotini and that helped review tweets with ambivalent labels. Additionally, we would like to express our sincere gratitude to the LightTag team and especially to Tal Perry for granting us free use for their annotation platform. 

\section*{Bibliographical References}
\label{main:ref}

\bibliographystyle{lrec}
\bibliography{lrec2020W-xample}

\end{document}